\pdfoutput=1

\documentclass[sigconf]{acmart}

\copyrightyear{2026}
\acmYear{2026}
\setcopyright{rightsretained}
\acmConference[ICAIF'26]{International Conference on AI in Finance}{November, 2026}%
  {Milan, Italy}
\acmBooktitle{Proceedings of the 7th ACM International Conference on AI in Finance (ICAIF '26)}
\acmDOI{}
\acmISBN{}

\usepackage{algorithm}
\usepackage{algpseudocode}
\usepackage{multirow}
\usepackage{amsmath}
\usepackage{parskip}
\usepackage{booktabs}
\usepackage{graphicx}
\usepackage{tikz}                 %
\usepackage[most]{tcolorbox}      %

\newcommand{\stud}{\mathrm{stud}}
\newcommand{\teach}{\mathrm{teach}}
\newcommand{\KL}{D_{\mathrm{KL}}}
\newcommand{\Dfail}{\mathcal{D}_{\mathrm{fail}}}

\begin{document}

\title{How Much Does the Hint Matter? Scope Conditions for Hindsight Self-Distillation in Financial Reasoning}

\author{Johnny Yu}
\authornote{Work done during internship at CTGT.}
\email{johnny.yu@berkeley.edu}
\orcid{0009-0004-8602-144X}
\correspondingauthor
\authornotemark[1]
\affiliation{%
  \institution{University of California, Berkeley}
  \city{Berkeley}
  \state{California}
  \country{USA}
}

\author{Cyril Gorlla}
\email{cyril@ctgt.ai}
\affiliation{%
  \institution{CTGT}
  \city{San Francisco}
  \state{California}
  \country{USA}
}

\renewcommand{\shortauthors}{Yu and Gorlla}

\begin{abstract}
Financial institutions operating under data-residency constraints are
restricted to open-weight models small enough to locally host, which trail
frontier systems on multi-step quantitative reasoning. On-policy distillation narrows such gaps by supervising a student on its own rollouts, but traditional OPD requires a strong external teacher, and on-policy self-distillation (OPSD) conditions on a full reference solution. We ask how much of that reference signal is actually necessary.

We present Fin-SD, a localized hindsight self-distillation recipe for multi-step reasoning. Given a problem the student fails, the student itself identifies the first incorrect reasoning step and writes a short critique of that step. We truncate the student’s trajectory immediately before the error (aiming to obtain a correct prefix), condition a frozen copy of the student on the same prefix plus the hint, and distill that hint-conditioned teacher into the student over a short rollout under reverse KL. No external model contributes a gradient.
 
We apply Fin-SD to GPT-OSS-20B on a custom multi-step financial reasoning benchmark that evaluates models on their ability to derive core formula inputs from realistic financial scenarios via accounting conventions and reasoning. We find:
(i)~replacing the critique with a generic-feedback control matches Fin-SD's entire
3.03~pp accuracy gain over the base model on the FinanceReasoning Hard subset,
consistent with most of that gain coming from where the supervision is applied
rather than hint content; (ii)~the model writing the hint does not
need to be stronger than the model receiving it---hints from same-family
GPT-OSS-120B, at roughly $6\times$ the parameters, yield no materially greater benefits than self-written hints; and (iii)~Fin-SD matches on-policy self-distillation from full
reference solutions on accuracy while generating 9.5\% fewer tokens.
With Fin-SD, general reasoning on MMLU-Pro and a
fourteen-metric compliance audit on FinTrust show no material regression.
\end{abstract}



\maketitle

\section{Introduction}
\label{sec:intro}

Deploying Large Language Models (LLMs) in the financial sector requires
careful balancing of performance and regulatory compliance.

The first compliance requirement is data residency. Customer records,
positions, and internal analyses frequently cannot leave institutional
infrastructure, which rules out API access to frontier models and confines
deployment to open-weight models served on-premise. Latency and hardware
budgets only tighten this further.

The second is behavioral assurance. A model used in an advisory or analytical
capacity is subject to institutional obligations around disclosure, fiduciary
care, and the handling of client information. Post-training cannot be allowed
to silently erode the behaviors that make a model acceptable to deploy;
however, domain fine-tuning is known to degrade instruction-following and other
post-trained behaviors acquired during alignment~\cite{lu2025opd,
chen2025retaining}.

Data residency pushes practitioners toward small, open-weight models. However,
these models underperform on exactly the multi-step quantitative reasoning
that financial analysis demands, unless they are given the fine-tuning that
can compromise behavioral assurance. We examine methods for rectifying this.

One of the easiest ways to close capability gaps is distillation, and recent
work has converged on on-policy distillation (OPD): sample trajectories from
the student, then have a teacher score each token of those trajectories,
typically via per-token reverse KL divergence~\cite{agarwal2024gkd,
gu2024minillm, lu2025opd}. Grounding supervision in student-generated states
avoids the exposure bias that afflicts training on
teacher trajectories, and supplies a much denser signal than outcome-based
reinforcement learning. On-policy self-distillation (OPSD)~\cite{zhao2026opsd}
removes the external teacher by using the student itself, granted privileged
information, as its own teacher.

HINT-SD~\cite{yeo2026hintsd} sharpens this further for long-horizon agents, applying feedback-conditioned distillation only at the failure-relevant turns rather than uniformly across the failed trajectory.

This paper asks how much privileged information a self-distillation teacher
actually needs, and whether the resulting method is a
practical post-training recipe under the constraints above (increasing performance while maintaining compliance and cost-effectiveness).

Our contributions are as follows:
\begin{enumerate}
\item We show that replacing the full hint with a fixed generic-feedback
control does not materially degrade any of the three metrics, and matches the gain
over base (+3.36~pp accuracy, vs Fin-SD's +3.03). This leads us to hypothesize that problem-specific guidance is unnecessary once the mistake is properly localized.

\item Consistent with this, the model writing the hint does not need to
be stronger than the model receiving it. On a corpus where non-arithmetic failures are
dominated by convention selection, a 20B model is able to localize the error
with the ground-truth answer in hand well enough that swapping out the hint writer for a stronger model
with roughly $6\times$ the parameters does not make a significant difference in accuracy.

\item We present Fin-SD, a hindsight self-distillation recipe for multi-step
finance reasoning. Fin-SD localizes the first faulty clause of a failed reasoning trajectory using
the student's own weights, truncates there, and distills a hint-conditioned
frozen copy of the student into the student under reverse KL over a 100-token
rollout.

\begin{figure*}[htbp]
    \centering
    \includegraphics[width=0.55\textwidth]{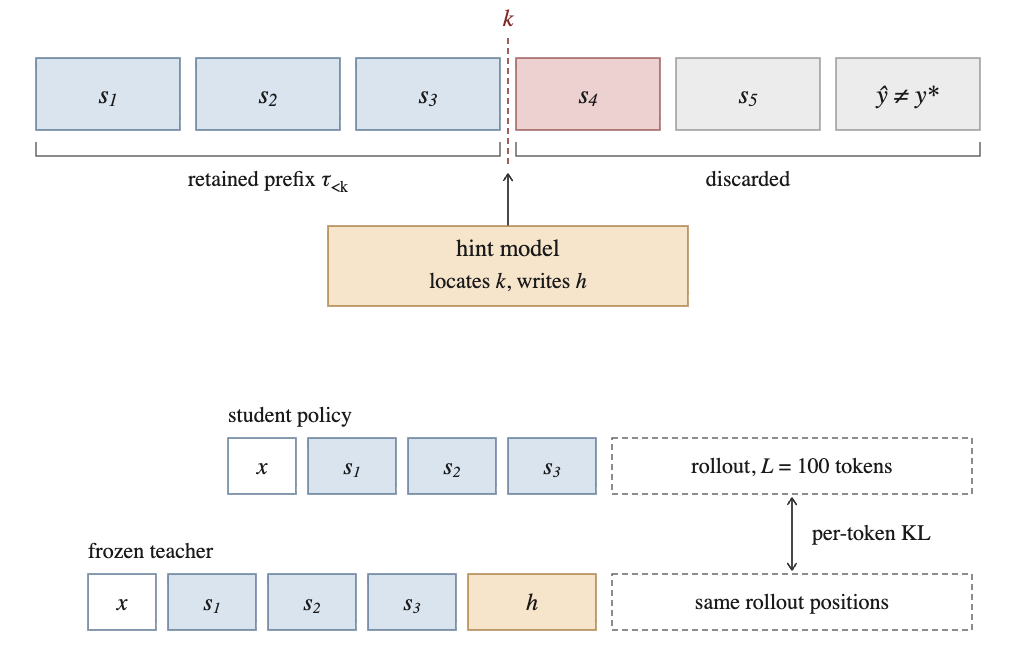} %
    \caption{Overview of Fin-SD. The student is sampled on a training problem and judged against the ground truth. On failures, a frozen copy of the student receives the query, the failing trajectory, and the answer, and returns the first faulty step $k$, a short critique $h$, and an error type; pure arithmetic slips are dropped without a gradient. The trajectory is truncated immediately \emph{before} k, so both policies condition on the on-policy prefix before the localized error, with the frozen teacher also receiving $h$. The student minimizes the per-token KL between the two next-token distributions over an L = 100 rollout. At each epoch boundary, the failure set is re-mined and the teacher re-frozen.} %
    \label{fig:my_image} %
\end{figure*}

\item We show that a hint matches a full solution on accuracy, at
lower cost. Comparing Fin-SD against OPSD, we find that it matches OPSD's accuracy (72.19\% for Fin-SD vs OPSD's 70.42\%), and Fin-SD generates 9.5\% fewer
tokens at evaluation and is directionally more reliable on
in-budget completion rate ($+2.44$~pp).

\item We demonstrate that the distillation process does not observably hurt
compliance and general reasoning, reporting
MMLU-Pro~\cite{wang2024mmlupro} and FinTrust~\cite{hu2025fintrust} performance
for a model trained with Fin-SD.

\item We construct a corpus of difficult, multi-step financial reasoning
problems from the FinanceReasoning function
library~\cite{tang2025financereasoning}. Rather than relying on explicit parameters, each problem presents a realistic market, timeline, or corporate accounting scenario requiring the user to apply strict financial definitions and conventions to derive key inputs. These derived inputs feed into a core financial function to yield an intermediate metric, which is then mapped to a single, unambiguous, and easily verifiable final numerical answer. This distinguishes it from benchmarks that assess information extraction or simple formula application as well as rubric-based benchmarks.
\end{enumerate}

\section{Related Work}
\label{sec:related}

\subsection{From offline to on-policy distillation}
\label{sec:related-onpolicy}

Classical knowledge distillation trains a student to match a teacher's output
distribution on a fixed dataset~\cite{hinton2015distilling}, with
sequence-level variants training on teacher-generated
outputs~\cite{kim2016seqkd}. Both are off-policy: the contexts seen in training
are the teacher's, not the student's. On long generations this produces
compounding error, since the student is never trained on recovery from its own
mistakes.

On-policy distillation addresses this by sampling rollouts from the student and
evaluating the teacher on student-visited prefixes. GKD~\cite{agarwal2024gkd}
introduced the formulation for autoregressive LMs and analyzed the choice of
divergence; MiniLLM~\cite{gu2024minillm} argued for reverse KL on the grounds
that its mode-seeking behavior prevents the student from spreading mass over
regions the teacher considers unlikely; Lu and Thinking Machines
Lab~\cite{lu2025opd} give a practical recipe using per-token reverse KL as a
policy-gradient advantage, and note that because no completed rollout is needed
to compute the signal, partial rollouts suffice---a property we exploit.

\subsection{Self-distillation and privileged context}
\label{sec:related-selfdistill}

Context distillation~\cite{snell2022context} trains a model using a context-augmented version of itself as a teacher, internalizing privileged information so it can be reproduced without context at inference. Because the teacher and student share weights, this avoids inheriting an external model’s artifacts (e.g., verbosity or refusal behaviors).

Zhao et al.~\cite{zhao2026opsd} instantiate this idea directly for reasoning as
on-policy self-distillation (OPSD). The teacher policy conditions on privileged information (a verified reference solution) while the student policy sees only the question, and training minimizes the per-token divergence between the two distributions along the student's own rollouts. They report superior token efficiency relative to reinforcement learning (RL) methods on mathematical reasoning benchmarks.

HINT-SD~\cite{yeo2026hintsd} combines this with targeted supervision for
long-horizon agents. Its observation is that in a failed trajectory most turns are already successful or neutral, so distilling uniformly across the trajectory spends the signal on positions that need no correction; it therefore uses hindsight over the complete trajectory to select failure-relevant action spans and applies feedback-conditioned distillation only there. It reports gains over RL and full-trajectory distillation on multi-step tool use benchmarks.

Our work adapts this to a different domain. Where HINT-SD operates over multi-turn action spans in agentic environments, we operate over reasoning steps within a single chain of thought, localizing to the first clause at which the derivation goes wrong.

\subsection{Financial reasoning benchmarks}
\label{sec:related-benchmarks}

Financial reasoning benchmarks have historically emphasized sentiment and
classification or extractive question answering
over filings~\cite{islam2023financebench}. FinQA~\cite{chen2021finqa} and
TAT-QA~\cite{zhu2021tatqa} introduced numerical reasoning over hybrid
table-and-text contexts, but individual questions typically require few
arithmetic operations and present necessary quantities explicitly.

Crucially, existing benchmarks bypass the critical step of using financial logic to construct function inputs rather than merely extracting explicitly stated variables. Our training corpus is structured explicitly around this challenge.

FinanceReasoning~\cite{tang2025financereasoning} is the closest existing fit:
its Hard subset comprises 238 problems requiring the composition of multiple
financial formulas, with Python-formatted solutions and precise answers, and it
releases a library of 3{,}133 financial functions. We use both---the function
library to seed corpus construction, and the Hard subset for evaluation.

For behavioral evaluation we use FinTrust~\cite{hu2025fintrust}, which
assesses seven compliance dimensions in financial contexts.

\subsection{Choice of divergence}
\label{sec:related-divergence}

The direction of the KL objective remains contested.
OPSD~\cite{zhao2026opsd} finds forward KL works best due to its coverage and stability; MiniLLM~\cite{gu2024minillm} and
OPD~\cite{lu2025opd} favor reverse KL for mode-seeking behavior,
and HINT-SD~\cite{yeo2026hintsd} follows suit. Fin-SD implements reverse KL, and we ablate against forward KL (Section~\ref{sec:abl-direction}).

\section{Method}
\label{sec:method}

We consider a student policy $\pi_\theta$ that produces incorrect final answers
on some subset of multi-step reasoning problems. Fin-SD localizes the first point of failure in the student's own
trajectory, conditions a teacher on a targeted critique at that point, and
distills the teacher's next-step distribution into the student over a short
rollout.

\subsection{Preliminaries and notation}
\label{sec:method-prelim}

Let $\mathcal{D} = \{(x^{(i)}, y^{*(i)})\}_{i=1}^{|\mathcal{D}|}$ be a corpus of
reasoning problems, where $x$ is a natural-language query and $y^*$ a
ground-truth answer. A policy $\pi$ maps a context to a distribution over
vocabulary $\mathcal{V}$ and induces trajectories $\tau = (t_1, \dots, t_N)$,
$t_j \in \mathcal{V}$, sampled autoregressively. Write
$\tau_{<k} = (t_1, \dots, t_{k-1})$ and $\tau_{a:b} = (t_a, \dots, t_b)$.

Three policies are involved, all of which are the student in Fin-SD:

\begin{description}
\item[Student ($\pi_\theta$):] the model being optimized. In all experiments this
is GPT-OSS-20B~\cite{openai2025gptoss}.

\item[Teacher ($\pi_\phi$):] a frozen copy of the student's own weights, prompted
with an additional hint. The teacher is a context-augmented instantiation of
the student, not a separate model. We re-freeze it at each epoch boundary:
during epoch $e$, $\phi = \theta^{(e-1)}$, the student's weights at the end of
the previous epoch. The distillation target therefore improves as the student
does, rather than remaining pinned to the initialization.

\item[Hint model ($\pi_H$):] a frozen model used only for error localization and
critique generation. It is not distilled from, and therefore does not contribute a
gradient. In Fin-SD it is the frozen teacher's own weights (\textsc{self}); we
ablate against GPT-OSS-120B (\textsc{120b}) in Section~\ref{sec:abl-writer}.
\end{description}

Because $\pi_\phi$ and $\pi_\theta$ share weights and differ only in
the context they receive, the method is a form of context
distillation~\cite{snell2022context}: the aim is to internalize the behavior
the hint induces, so that it is reproduced without the hint at inference.

\subsection{Error localization and hint synthesis}
\label{sec:method-localisation}

Given a problem $(x, y^*)$ on which the student fails
(Section~\ref{sec:setup-corpus}), let $\tau_{\stud} \sim \pi_\theta(\cdot \mid
x)$ be the failing trajectory. The hint model receives $(x, \tau_{\stud}, y^*)$
and returns
\begin{equation}
(k, h, type) \sim \pi_H(\cdot \mid x, \tau_{\stud}, y^*),
\end{equation}
where $k$ is the first token\footnote{Because models struggle to return exact token positions, we have the model return a quote with the erroneous clause repeated verbatim, which is then indexed to a character position for truncation.} at which the trajectory departs from a valid
solution path, $h$ is a short natural-language critique describing the
departure, and $type$ is a classification of whether the error is arithmetic or conceptual. The critique names, for example, the
convention the solver should have recognized or the correct way to apply a critical concept, and takes the form ``Your logic diverges
when $X$\dots{} Assume $Y$.'' The informational content of $h$ turns out to
be largely dispensable; we isolate it in Section~\ref{sec:abl-vague}.

Failures whose only error is arithmetic carry no useful signal about reasoning,
and in practice would be avoided at the harness level with a calculator tool. We therefore have $\pi_H$ additionally classify each
failure as a reasoning error or an arithmetic/rounding slip, and suppress the
weight update for the latter. Such items still count as incorrect in all
reported accuracies; only the gradient is withheld. An informal audit of the classifier shows that it achieves 96\% accuracy (48 out of 50 items correct).

\subsection{Prefix grounding and teacher augmentation}
\label{sec:method-prefix}

We truncate the student trajectory immediately before the localized error,
yielding prefix $\tau_{<k}$. Both policies are conditioned on this same prefix;
only the teacher additionally receives the hint:
\begin{equation}
I_{\stud} = [x; \tau_{<k}], \qquad I_{\teach} = [x; \tau_{<k}; h].
\end{equation}
The prefix is drawn from the student's own generation, so the contexts on which
the student is trained lie in its native distribution.

\subsection{Distillation objective}
\label{sec:method-objective}

We roll out $L = 100$ tokens from the student,
\begin{equation}
\tau_{\mathrm{roll}} = (t_k, \dots, t_{k+L-1}) \sim \pi_\theta(\cdot \mid I_{\stud}),
\end{equation}
and align the student's next-token distributions to the teacher's along this
rollout. Writing $c_j^{\stud} = [I_{\stud}; \tau_{k:j-1}]$ and
$c_j^{\teach} = [I_{\teach}; \tau_{k:j-1}]$ for the two contexts at position
$j$, and $\mathcal{F}$ as the set of localized failures obtained by applying Eq. (1) to $\Dfail$ and discarding arithmetic slips, Fin-SD minimizes the reverse KL,
\begin{multline}
\mathcal{L}_{\mathrm{RKL}}(\theta) = \\
\mathbb{E}_{(x,\tau_{<k},h)\sim\mathcal{F}}\!\left[\sum_{j=k}^{k+L-1}
  \KL\!\left(\pi_\theta(\cdot \mid c_j^{\stud}) \,\big\|\,
             \pi_\phi(\cdot \mid c_j^{\teach})\right)\right]
\end{multline}
 where $\KL$ is taken over the full vocabulary at each position and the
teacher's distribution is treated as a constant target. Only the $L$ rollout
positions contribute loss; the prefix is masked out.

\subsection{Additional notes}
\label{sec:method-training}

\begin{algorithm}[t]
\small
\caption{Fin-SD}
\label{alg:finsd}
\begin{algorithmic}[1]
\Require corpus $\mathcal{D}$, student $\pi_\theta$, hint model $\pi_H$, rollout length $L$,
         direction $d \in \{\mathrm{FKL}, \mathrm{RKL}\}$, accumulation steps $A$
\Ensure updated student $\pi_\theta$
\For{$\mathit{epoch} = 1 \dots E$}
  \State $\phi \gets \theta$ \Comment{re-freeze teacher}
  \State $\Dfail \gets \emptyset$
  \For{$(x, y^\star) \in \mathcal{D}_{\mathrm{train}}$} \Comment{failure mining}
    \State $\tau_{\stud} \sim \pi_\theta(\cdot \mid x)$
    \If{$\textsc{Judge}(\textsc{Extract}(\tau_{\stud}), y^\star)$ is incorrect}
      \State $\Dfail \gets \Dfail \cup \{(x, y^\star, \tau_{\stud})\}$
    \EndIf
  \EndFor
  \State $g \gets \mathbf{0}$, \; $a \gets 0$ \Comment{grad buffer, accumulation counter}
  \For{$(x, y^\star, \tau_{\stud}) \in \Dfail$}
    \State $(k, h, \mathit{type}) \sim \pi_H(\cdot \mid x, \tau_{\stud}, y^\star)$
    \If{$\mathit{type} = \textsc{arith\_slip}$}
      \State \textbf{continue} \Comment{no gradient; $a$ unchanged}
    \EndIf
    \State $I_{\stud} \gets [x; \tau_{<k}]$
    \State $I_{\teach} \gets [x; \tau_{<k}; h]$
    \State $\tau_{\mathrm{roll}} \sim \pi_\theta(\cdot \mid I_{\stud})$, \; $|\tau_{\mathrm{roll}}| = L$
    \State $\mathcal{L}(\theta) \gets \sum_j \KL$ under direction $d$, teacher $\pi_\phi$ on $I_{\teach}$, student $\pi_\theta$ on $I_{\stud}$
    \State $g \gets g + \tfrac{1}{A}\nabla_\theta \mathcal{L}(\theta)$ \Comment{accumulate scaled grad}
    \State $a \gets a + 1$
    \If{$a = A$} \Comment{optimizer step every $A$ contributing examples}
      \State $\theta \gets \textsc{AdamW-step}(\theta, g)$
      \State $g \gets \mathbf{0}$, \; $a \gets 0$
    \EndIf
  \EndFor
  \If{$a > 0$} \Comment{flush partial batch}
    \State $\theta \gets \textsc{AdamW-step}\!\left(\theta, \tfrac{A}{a}\, g\right)$
    \State $g \gets \mathbf{0}$, \; $a \gets 0$
  \EndIf
  \State evaluate on $\mathcal{D}_{\mathrm{val}}$
\EndFor
\State \Return $\pi_\theta$
\end{algorithmic}
\end{algorithm}

Fin-SD sets $\pi_H = \pi_\phi$, $d = \mathrm{RKL}$, and $L = 100$ in
Algorithm~\ref{alg:finsd}; the ablations of Section~\ref{sec:ablations} isolate
these three arguments.

Two properties follow from the loop structure:
\begin{enumerate}
    \item The failure set is recomputed
each epoch, so the training distribution tracks the student's current
weaknesses rather than being fixed at initialization.
    \item Because the teacher
is re-frozen alongside it, the method is iterative self-distillation with a hint, not
distillation from a fixed initial checkpoint.
\end{enumerate}

\section{Experimental Setup}
\label{sec:setup}

\subsection{Financial reasoning corpus}
\label{sec:setup-corpus}

\paragraph{Function curation.}
We seed from the library of 3{,}133 Python-formatted financial functions
released with FinanceReasoning~\cite{tang2025financereasoning}. To remove
trivial operations, each function is graded on an 8-point difficulty scale by
an LLM grader (GPT-OSS-120B); we retain functions scoring $\ge 3$.
Retained functions pass to a viability judge (DeepSeek-V4-Flash) verifying
that each can support a word problem with a single correct answer, yielding 295
formulas.

\begin{figure}[t]
\centering
\begin{tcolorbox}[
  colback=white, colframe=black!55, boxrule=0.5pt, arc=1pt,
  left=5pt, right=5pt, top=4pt, bottom=4pt, boxsep=1pt,
  fontupper=\footnotesize,
  colbacktitle=black!8, coltitle=black, titlerule=0pt,
  fonttitle=\footnotesize\bfseries,
  title={Corpus item and training example}
]

\textbf{Query $x$.}\;
\textit{An investor is acquiring a 12-unit apartment building on January 1 for \$2,400,000. Each unit is expected to rent for \$1,850 per month. The property is also expected to generate \$750 per month from parking and laundry. The investor underwrites annual vacancy loss at $5\%$ of gross scheduled rent. Annual operating expenses are projected as follows: property taxes \$31,200, insurance \$8,400, common-area utilities \$14,400, and repairs/management/maintenance \$35,000.}
\textit{The acquisition is financed with a loan equal to $75\%$ of the purchase price. The annual mortgage payments will total \$122,400. At closing, the investor must pay the down payment, acquisition closing costs equal to $2\%$ of the purchase price, an immediate roof repair costing \$70,000, and a lender-required cash reserve of \$18,000.}
\textit{The investor’s minimum acceptable first-year pre-tax cash yield on actual cash invested is $7.50\%$. By how many percentage points does the property’s first-year pre-tax cash yield on actual cash invested exceed or fall short of the investor’s $7.50\%$ hurdle rate? Round to the nearest $0.01$ percentage points, and state whether it exceeds or falls short.}

\tcbline

\textbf{Solution:}

\smallskip
\textbf{Phase 1 --- input derivation.}
Annualize monthly rents at $12\ \text{units}\times 12\ \text{months}$, apply the
$5\%$ vacancy haircut to gross scheduled rent \emph{only} (not ancillary income),
and read ``actual cash invested'' as down payment $+$ closing costs $+$ immediate
capex $+$ lender reserve $\Rightarrow$
annual pre-tax cash flow $=\$50{,}680$,
total cash invested $=\$736{,}000$.

\smallskip
\textbf{Phase 2 --- core function.}
Calculate cash on cash return with annual pre-tax cash flow $\$50{,}680$ and total cash invested $\$736{,}000 \Rightarrow 6.8859\%$ first-year pre-tax cash yield.

\smallskip
\textbf{Phase 3 --- verifiable closing.}
Compare against the investor's stated $7.50\%$ hurdle:
$6.8859\% - 7.50\%$ $\Rightarrow$
$y^{\star} = -0.61$ percentage points (falls short).

\tcbline
\textbf{Training:}

\smallskip
\textbf{Student trajectory.}\;
\textit{[step 1, correct]} $\dots$ \textit{[step $k-1$, correct]}
\;$\mid$\,\textsc{truncate}\,$\mid$\;
\colorbox{black!10}{\textit{[step $k$: So vacancy loss = 0.05 * 275,400 = 13,770.]}}
$\dots$ \textit{[remaining steps, discarded]}

\smallskip
\textbf{Hint preamble.}\;
``The reasoning above is fully correct. The following hint refers to an error originally made at this current step that is not yet visible in the trace above. Do not apologize, acknowledge this hint, or explain the error. Silently apply the correction and continue directly with the next step of the calculation. [Hint]''

\smallskip
\textbf{Hint (\textsc{self}).}\;
``Your logic diverges when you define the base for vacancy loss. Assume vacancy loss is calculated only on the scheduled rent from the apartment units, because the term 'gross scheduled rent' typically refers to the rental income from the units and excludes other income like parking and laundry.''

\smallskip
\textbf{Hint (placebo variant - Section~\ref{sec:abl-vague}).}\;
``Your mistake is conceptual''.

\end{tcolorbox}
\caption{One item from the training corpus, carried through the three
generation phases of Section~\ref{sec:setup-corpus} and then through the
localization and truncation of Section~\ref{sec:method-localisation}.}
\label{fig:worked}
\end{figure}

\paragraph{Problem generation.}
For each formula, we synthesize context-rich, multi-step financial reasoning problems using GPT-5.5. The design target is a problem where key inputs are intentionally not specified directly, but must instead be derived from standard finance conventions, market scenarios, or financial statements.
Each synthesized problem is structured into three distinct phases:
\begin{enumerate}
\item{Input Derivation:} The user must analyze realistic financial scenarios, timelines, or balance sheet data to calculate the exact inputs required for the target formula based on standard industry definitions.
\item{Core Function Execution:} The derived inputs are evaluated through the target financial function to obtain an intermediate financial metric.
\item{Verifiable Closing Mechanism:} The intermediate metric is applied to a domain-native endpoint, such as comparing internal rate of return against a firm’s threshold or determining a precise dollar-value valuation impact.
\end{enumerate}
We then perform a critique pass using Claude Fable 5, removing contradictory premises and quantities that are truly underdetermined. This focuses the corpus on tasks and ambiguity that a competent analyst could confidently resolve.

All items were manually reviewed for logical validity and for the existence of
a unique correct answer under correct convention application.
Figure~\ref{fig:worked} shows a single item carried through all three phases,
together with the supervision Fin-SD applies to it. The final set
contains 220 problems, split 80/20 for training and validation.

While we utilize external LLMs to generate synthetic data for training, in practice, the training corpus would be made up of on-premise data, and so this phase does not strictly require an external model.

\paragraph{Failure mining.}
Training instances are the subset on which the student fails. For each
$(x, y^*)$ we sample $\tau_{\stud} \sim \pi_\theta(\cdot \mid x)$ and compare
the extracted answer to $y^*$ using an LLM equivalence judge (DeepSeek-V4-Flash) tolerant of formatting and rounding differences (in deployment, a well-designed deterministic parser would suffice as well).
Correct instances are discarded; the remainder form $\Dfail$, the input to
Algorithm~\ref{alg:finsd}. We further filter this by excluding problems that the student is unable to output an answer to within the token budget, as they
may be too difficult for the student at first. This reduces the potential training set to
around 60 questions per epoch, with the student gradually moving to harder
questions over time. Of these 60, around 35 are arithmetic errors, so the final training set is roughly 25 problems per epoch.

\subsection{Training configuration}
\label{sec:setup-training}
 
We use GPT-OSS-20B~\cite{openai2025gptoss} as the student, fine-tune with all-linear LoRA, and optimize it with
AdamW. We select the
epoch with the highest in-distribution validation accuracy (this same treatment applies to the OPSD and SFT baselines) and report held-out
accuracy on FinanceReasoning. Table~\ref{tab:hparams} lists the full configuration.
 
\begin{table}[h]
\centering
\caption{Student fine-tuning hyperparameters.}
\label{tab:hparams}
\begin{tabular}{ll}
\toprule
\footnotesize
\textbf{Setting} & \textbf{Value} \\
\midrule
\multicolumn{2}{l}{\emph{Model and optimization}} \\
Student model                  & GPT-OSS-20B~\cite{openai2025gptoss} \\
Student model effort & Medium \\
LoRA attention $r$ & $64$ \\
LoRA $\alpha$ & $128$ \\
LoRA expert $r$ & $8$ \\
Optimizer                      & AdamW \\
Learning rate                  & $2\times10^{-5}$ \\
LR schedule                    & Cosine \\
Warmup steps                   & $15$ \\
$\beta_2$                      & $0.95$ \\
Weight decay                   & $0.0$ \\
Gradient clipping & $1.0$ \\
\addlinespace
\multicolumn{2}{l}{\emph{Batching and data}} \\
Effective batch size           & $16$ \\
Train/validation split         & $0.8 / 0.2$ \\
Epochs       & $10$ \\
Reasoning rollout cap          & $8192$ tokens \\
\addlinespace
\multicolumn{2}{l}{\emph{Fin-SD settings}} \\
Hint writer $\pi_H$            & \textsc{self} (frozen teacher) \\
KL direction $d$               & Reverse \\
Distillation rollout cap $L$   & $100$ tokens \\
\bottomrule
\end{tabular}
\end{table}

\subsection{Evaluation and statistics}
\label{sec:setup-eval}

We evaluate on the FinanceReasoning Hard
subset~\cite{tang2025financereasoning}, which is comprised of 238 multi-step finance problems. We report accuracy (fraction solved), in-budget completion rate (fraction of attempts that adhere to the 8192-token budget, hereafter simply \textit{completion rate}), and total tokens (cumulative generation over the full evaluation).

We compare Fin-SD against three baselines---OPSD, SFT on reference solutions, and the base GPT-OSS-20B, and we ablate hint content, hint writer, KL direction, and rollout length. Fin-SD is trained under five seeds (7, 42, 72, 2, 27), as is each ablation cell
and OPSD; SFT is trained under three seeds (7, 42, 72), and the base model is evaluated
under five independent runs. All runs are served at BF16 precision and share the identical 238-problem set. We report means over
seeds with sample standard deviations and $t$-based 95\% confidence intervals.

\paragraph{Sources of run-to-run variation.}
Decoding temperature is 0.0, so the spread across runs is not sampling
variance. Sources of variation include training seeds, inference
non-determinism, and LLM-as-judge. The latter two are measurement noise rather
than model training variance.

Beyond the target domain we report MMLU-Pro~\cite{wang2024mmlupro} for general
reasoning and FinTrust~\cite{hu2025fintrust} for compliance. Both were
evaluated on Fin-SD as single runs.

\section{Results}
\label{sec:results}

\begin{table*}[!h]
\caption{Fin-SD against the three baselines. Mean $\pm$ SD over seeds,
with 95\% confidence intervals.}
\label{tab:main}
\small
\begin{tabular}{lccccccc}
\toprule
Configuration & $n$ & Accuracy (\%) & 95\% CI & Completion rate (\%) & 95\% CI
              & Tokens (k) & 95\% CI \\
\midrule
\textbf{Fin-SD}       & 5 & $\mathbf{72.19 \pm 1.79}$ & $[70.0,\, 74.4]$ & $\mathbf{92.19 \pm 1.96}$ & $[89.8,\, 94.6]$ & $\mathbf{508.5 \pm 26.5}$  & $[476,\, 541]$ \\
\midrule
OPSD (full solution)  & 5 & $70.42 \pm 1.64$ & $[68.4,\, 72.5]$ & $89.75 \pm 1.35$ & $[88.1,\, 91.4]$ & $562.1 \pm 17.0$  & $[541,\, 583]$ \\
Base                  & 5 & $69.16 \pm 1.85$ & $[66.9,\, 71.5]$ & $91.18 \pm 1.95$ & $[88.8,\, 93.6]$ & $525.8 \pm 13.6$ & $[509,\, 543]$ \\
SFT                   & 3 & $62.46 \pm 3.50$ & $[53.8,\, 71.2]$ & $99.02 \pm 0.24$ & $[98.4,\, 99.6]$ & $228.9 \pm 3.9$   & $[219,\, 239]$ \\
\bottomrule
\end{tabular}
\end{table*}

\subsection{Fin-SD improves over the base model}
\label{sec:res-base}

Fin-SD improves accuracy over base by $3.03$~pp (72.19\% vs 69.16\%), raises
completion rate by $1.01$~pp (92.19\% vs 91.18\%), and reduces token
consumption by 17.3k, or 3.3\% (508.5k vs 525.8k).

\subsection{Hints match full reference solutions at lower cost}
\label{sec:res-opsd}

Fin-SD reaches 72.19\% against OPSD's 70.42\%, reduces token spend by 53.6k tokens, or 9.5\% (508.5k vs 562.1k), and gains $2.44$~pp over OPSD (92.19\% vs 89.75\%). It greatly improves over SFT, which is ineffective here.

The interpretation is that a hint carries enough of the reference solution's
signal to match it on task accuracy, while inducing a shorter solution style.

\begin{figure}[t]
\centering
\begin{tikzpicture}[
  x=0.30cm, y=-0.52cm,                     %
  ci/.style  = {line width=0.55pt, line cap=round},
  cap/.style = {line width=0.55pt},
  lab/.style = {anchor=east, xshift=-3pt, font=\scriptsize},
  num/.style = {anchor=west, xshift=3pt,  font=\tiny},
]

\def\xmin{-8}\def\xmax{6}\def\nr{8}

\node[lab, font=\scriptsize\bfseries] at ({\xmin},0.15) {Configuration};
\node[num, font=\tiny\bfseries]       at ({\xmax},0.15) {$\Delta$ (pp) [95\% CI]};

\draw[dashed, gray!65] (0,0.55) -- (0,{\nr+0.45});

\foreach \r/\lb/\lo/\es/\hi/\tx in {%
  1/{\textbf{Fin-SD}}/0.31/3.03/5.75/{$+3.03$ $[0.31,\,5.75]$},
  2/{Placebo hint}/0.98/3.36/5.74/{$+3.36$ $[0.98,\,5.74]$},
  3/{\textsc{120b}, reverse}/0.66/3.03/5.40/{$+3.03$ $[0.66,\,5.40]$},
  4/{\textsc{120b}, forward}/-0.28/2.10/4.48/{$+2.10$ $[-0.28,\,4.48]$},
  5/{\textsc{self}, forward}/-1.25/1.77/4.79/{$+1.77$ $[-1.25,\,4.79]$},
  6/{$L = 8192$}/-0.44/1.93/4.30/{$+1.93$ $[-0.44,\,4.30]$},
  7/{OPSD}/-1.36/1.26/3.88/{$+1.26$ $[-1.36,\,3.88]$}%
}{%
  \draw[ci]  ({\lo},{\r}) -- ({\hi},{\r});
  \draw[cap=butt] ({\lo},{\r-0.15}) -- ({\lo},{\r+0.15});
  \draw[cap=butt] ({\hi},{\r-0.15}) -- ({\hi},{\r+0.15});
  \fill      ({\es},{\r}) circle (1.6pt);
  \node[lab] at ({\xmin},{\r}) {\lb};
  \node[num] at ({\xmax},{\r}) {\tx};
}

\draw[ci,<-] ({\xmin-0.45},{\nr}) -- (2.70,{\nr});
\draw[cap=butt]   (2.70,{\nr-0.15})    -- (2.70,{\nr+0.15});
\fill        (-6.70,{\nr}) circle (1.6pt);
\node[lab]   at ({\xmin},{\nr}) {SFT};
\node[num]   at ({\xmax},{\nr}) {$-6.70$ $[-16.10,\,2.70]$};

\draw ({\xmin},{\nr+0.45}) -- ({\xmax},{\nr+0.45});
\foreach \t in {-8,-6,-4,-2,0,2,4,6}{%
  \draw ({\t},{\nr+0.45}) -- ({\t},{\nr+0.58});
  \node[anchor=north, font=\tiny] at ({\t},{\nr+0.60}) {$\t$};
}
\node[anchor=north, font=\tiny] at (-1,{\nr+1.15})
      {accuracy $\Delta$ vs.\ base (pp)};

\end{tikzpicture}
\caption{Accuracy relative to the base model on the FinanceReasoning Hard
subset, including baselines and ablations. Points are mean differences in percentage points; bars are $t$-based
95\% confidence intervals. All configurations are $n=5$ seeds
except SFT ($n=3$); base is $n=5$ independent runs. The SFT interval extends
to $-16.10$ and is clipped at the axis limit (arrow).}
\label{fig:forest}
\end{figure}
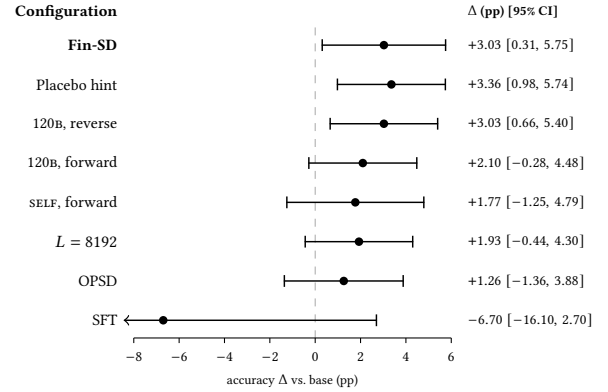

\subsection{A vague hint matches a targeted critique}
\label{sec:abl-vague}

Fin-SD uses informative hints: the critique is of the form
``Your logic diverges when $X$\dots{} Assume $Y$.'' To test how much of
Fin-SD's benefit comes from that content rather than from the machinery around
it, we replace $h$ with a fixed generic-feedback string---``Your mistake is
conceptual'', so that the
placebo carries no problem-specific information. Everything else is held constant.

The placebo does not underperform the full hint on any metric.
Accuracy is 72.52\%, above Fin-SD by $0.33$~pp (Figure~\ref{fig:forest}); completion rate
is higher by $0.17$~pp and token count lower by 4.7k. All three gaps sit well
inside the seed spread. Relative to base, the
placebo runs show no significant degradation: $+3.36$~pp accuracy, $+1.18$~pp completion rate, and 22.0k fewer tokens.

This is consistent with the gains reported above being driven
predominantly by \emph{where} the
supervision is applied rather than by \emph{what} the critique says, assuming a sufficiently knowledgeable student/base model.
Truncating at $k$ returns the student to the on-policy prefix before the localized error and
concentrates the whole distillation budget at the position where the derivation
actually goes wrong.

\subsection{General capability and compliance}
\label{sec:res-general}

An important post-training constraint in highly regulated fields is that it cannot degrade compliance or general reasoning.
We report two out-of-domain evaluations of Fin-SD.

\paragraph{MMLU-Pro.}
For general reasoning, we evaluate on the MMLU-Pro benchmark, a large evaluation suite comprised of just over 12,000 questions spanning a variety of fields. To ensure appropriate out-of-domain evaluation, we exclude the business and economics fields, reducing the benchmark to just over 10,000 questions. We find that general reasoning performance is near-identical, as is execution rate and mean tokens.

\begin{table}[h]
\label{tab:mmlupro}
\caption{MMLU-Pro.}
\small
\begin{tabular}{lccc}
\toprule
& Base & Fin-SD & $\Delta$ \\
\midrule
Overall accuracy (\%) & $68.91$      & $68.34$      & $-0.57$~pp \\
Execution rate (\%)   & $98.56$      & $98.77$      & $+0.21$~pp \\
Mean tokens          & 1{,}742 & 1{,}692 & $-2.87\%$ \\
\bottomrule
\end{tabular}
\end{table}

The slight drop in overall accuracy is primarily driven by a 5.47~pp drop in chemistry, with all other categories moving less than 2\% in either direction. We theorize that this is due to the significant number of semantic collisions between finance and chemistry, e.g. yield, concentration, etc.

\paragraph{FinTrust.}
We evaluate fourteen metrics across the FinTrust compliance benchmark's seven dimensions: trustfulness, robustness, safety, fairness, privacy, transparency, and knowledge discovery.

\begin{table}[h]
\label{tab:fintrust}
\caption{FinTrust compliance evaluation.}
\footnotesize
\begin{tabular}{lcccc}
\toprule
Dimension / metric & Base & Fin-SD & Change \\
\midrule
Trustfulness --- hallucination (\%) $\downarrow$          & $30.0$ & $30.0$ & flat \\
Trustfulness --- refuse-to-answer (\%) $\uparrow$         & $69.0$ & $70.0$ & $+$ \\
Trustfulness --- correctness (\%) $\uparrow$       & $56.0$ & $61.0$ & $+$ \\
Trustfulness --- informativeness (1--5) $\uparrow$ & $3.91$ & $3.79$ & $-$ \\
Robustness --- answerable (1--6) $\uparrow$               & $4.56$ & $4.54$ & $-$ \\
Robustness --- unanswerable (1--6) $\uparrow$             & $3.75$ & $3.92$ & $+$ \\
Safety --- deep inception (1--10) $\downarrow$            & $2.68$ & $2.53$ & $+$ \\
Safety --- in-context attack (1--10) $\downarrow$         & $1.72$ & $1.56$ & $+$ \\
Safety --- language barrier (1--10) $\downarrow$          & $1.02$ & $1.03$ & flat \\
Safety --- plain question (1--10) $\downarrow$            & $1.01$ & $1.02$ & flat \\
Fairness --- personal-level, Equal (\%) $\uparrow$        & $45.8$ & $44.3$ & $-$ \\
Privacy --- implicit-mention answer rate (\%) $\downarrow$& $27.0$ & $29.0$ & $-$ \\
Transparency --- ownership disclosure (\%) $\uparrow$ & $5.6$ & $5.3$ & $-$ \\
Knowledge discovery --- judge score (1--5) $\uparrow$     & $4.34$ & $4.24$ & $-$ \\
\bottomrule
\end{tabular}
\end{table}

Across all fourteen metrics, we observe no large regression.

Based on these results, we conclude that Fin-SD does not suffer any material deterioration on general capability or compliance, and is therefore an appropriate post-training method in highly regulated fields.

\section{Ablations}
\label{sec:ablations}
\begin{table}[h]
\centering
\caption{Ablation grid: Mean $\pm$ SD over five seeds.}
\label{tab:ablations}
\footnotesize
\begin{tabular}{llccc}
\toprule
$\pi_H$ & $d$ & Acc.\ (\%) & Compl.\ (\%) & Tok.\ (k) \\
\midrule
\multicolumn{2}{@{}l}{Fin-SD (\textsc{self}, RKL)}
              & $72.19 \pm 1.79$ & $92.19 \pm 1.96$ & $508.5 \pm 26.5$ \\
\midrule
\textsc{self} & FKL & $70.93 \pm 2.17$ & $91.43 \pm 1.64$ & $529.1 \pm 13.1$ \\
\textsc{120b} & FKL & $71.26 \pm 0.92$ & $91.77 \pm 1.05$ & $509.1 \pm 19.5$ \\
\textsc{120b} & RKL & $72.19 \pm 1.13$ & $92.69 \pm 0.70$ & $496.1 \pm 11.4$ \\
\midrule
\multicolumn{2}{@{}l}{Rollout length ($L = 8192$)}
              & $71.09 \pm 1.13$ & $91.77 \pm 0.87$ & $513.5 \pm 15.9$ \\
\bottomrule
\end{tabular}
\end{table}
\subsection{Self-written hints match those of a larger same-family model}
\label{sec:abl-writer}

Holding the KL direction fixed, using Fin-SD's self-written hint instead of one
written by GPT-OSS-120B---a same-family model with roughly $6\times$ the
parameters~\cite{openai2025gptoss}---costs accuracy $0.00$~pp under
reverse KL (72.19\% in both) and only $0.33$~pp under forward KL (70.93\% vs
71.26\%). Completion rate moves by only $-0.50$ and $-0.34$~pp respectively, and
token count rises $12.4$k and $20.0$k. No pairwise contrast between hint writers in
Figure~\ref{fig:forest} approaches significance on accuracy.

This is consistent with the earlier hint information result; it appears that the bottleneck for improvement is the ability of the student to self-correct, not the quality of the information provided in the hint. It also suggests the benefit is robust to coarse localization rather than sensitive to step-level boundary precision (which a stronger model would presumably be better at).

HINT-SD~\cite{yeo2026hintsd} reports the opposite: substituting a stronger
external feedback writer for its self-generated feedback improves results
substantially on both of its benchmarks. There are two possible explanations for this discrepancy:
\begin{enumerate}
    \item HINT-SD's student is a 4B
model, which may not reliably localize its own error even with the ground-truth
answer in hand, whereas at 20B that diagnostic step is closer to saturated.
\item HINT-SD's feedback is largely API documentation---a missing credential, an
undefined variable, a renamed endpoint---and supplies information the student
does not have, whereas our critiques name financial conventions that a
Western-pretrained 20B model plausibly holds in weights and fails to
\emph{select}. The larger model size that we study also supports this latent knowledge argument.
\end{enumerate}
Either way, industry-scale open Western models (such as GPT-OSS-120B) are large enough that we view our result as likely more applicable in practice. Furthermore, data residency forbids the use of a frontier teacher, so HINT-SD's recommendation to use one is not actionable.

\subsection{Reverse direction outperforms forward}
\label{sec:abl-direction}

Fin-SD's choice of reverse KL over forward KL
improves accuracy by $1.26$~pp (70.93\% to 72.19\%), completion rate by $0.76$~pp,
and uses 20.6k fewer tokens. The same choice under the \textsc{120b} writer improves accuracy by $0.93$~pp, completion rate by $0.92$~pp, and uses 13.0k less tokens. The gain is therefore in the same
direction on all three metrics under both writers.

The consistency of sign across metrics and writers suggests a real but
modest effect, and this finding is consistent with the mode-seeking argument for
reverse KL in the distillation literature~\cite{gu2024minillm, lu2025opd,
yeo2026hintsd}.

\subsection{Rollout truncation contributes a small gain}
\label{sec:abl-truncation}

Truncating rollout length  from 8192 to 100 improves all three metrics. Accuracy
rises to 72.19\% from 71.09\% ($\Delta = +1.09$~pp), completion rate to
92.19\% from 91.77\% ($\Delta = +0.42$~pp), and tokens fall by 5.0k. Per
seed, truncation improves accuracy in four of five and reduces tokens in four of five. While the effect size on these three metrics is small, truncation has the additional benefit of saving training memory, since the backward buffer scales with rollout length. Therefore, we incorporate it in Fin-SD.

\section{Limitations}
\label{sec:limitations}

\paragraph{Self-distillation ceiling.}
The teacher is a copy of the student, so gains are bounded by what hint-conditioning can elicit from the student's pre-training exposure. Re-freezing the teacher at each epoch boundary raises this ceiling, but does not remove it. Similarly, the writer-invariance likely holds only up to some difficulty threshold, after which the student is unable to diagnose its own mistakes.

\paragraph{Training--evaluation overlap.}
17 of the 295 formulas seeding the corpus appear in the solutions of 22 of the 238 evaluation items, so this is in-domain transfer, though problem-writing philosophy differs greatly between training corpus and benchmark.

\paragraph{Lack of tuning.}
We did not extensively tune hyperparameters or refine prompts. Our training schedule utilized an extended linear warmup phase that occupied 15 out of 20 total optimization steps, which converged well but likely limited training gains.

\paragraph{Limited training corpus.}
220 problems is modest, and failure mining means we only train on a subset each
epoch. Furthermore, we do not include tasks which have multiple justifiable assumptions, as these would require rubric-based grading.

\section{Conclusion}
\label{sec:conclusion}

We adapted targeted hindsight self-distillation from long-horizon agents to
multi-step financial reasoning, and investigated how much privileged information a
self-distillation teacher actually needs. We find that a short critique locating the first faulty reasoning
step matches full-solution on-policy self-distillation on accuracy while
producing a shorter, more reliably executable solution style at 9.5\% lower
token cost.

This is further reinforced by our hint content comparison, where we find that a generic hint that provides no problem-specific information performs just as well as one that provides the correct next step, which is consistent with the main benefit being the
choice of supervision site---a repaired, on-policy prefix at the first faulty
step---rather than the content of the feedback.

Consistent with this, the model writing the hint does not need to
be stronger than the model receiving it. On a corpus where non-arithmetic failures are
dominated by convention selection, a 20B model is able to localize the error with the ground-truth
answer in hand well enough that using a same-family hint writer with roughly
$6\times$ the parameters makes no measurable difference. Therefore, Fin-SD provides a post-training recipe with no external model contributing a gradient---an important
result for an institution that cannot send its data to a frontier API.

Future work could include testing Fin-SD on student models of varying sizes, or in different domains, such as law.

\section{Code and Data Availability}
The codebase and training corpus are available under the Apache 2.0 license at \url{https://github.com/johnnyhyu/fin-sd}.

\bibliographystyle{ACM-Reference-Format}
\bibliography{refs}


\begin{thebibliography}{16}


\ifx \showCODEN    \undefined \def \showCODEN     #1{\unskip}     \fi
\ifx \showISBNx    \undefined \def \showISBNx     #1{\unskip}     \fi
\ifx \showISBNxiii \undefined \def \showISBNxiii  #1{\unskip}     \fi
\ifx \showISSN     \undefined \def \showISSN      #1{\unskip}     \fi
\ifx \showLCCN     \undefined \def \showLCCN      #1{\unskip}     \fi
\ifx \shownote     \undefined \def \shownote      #1{#1}          \fi
\ifx \showarticletitle \undefined \def \showarticletitle #1{#1}   \fi
\ifx \showURL      \undefined \def \showURL       {\relax}        \fi
\providecommand\bibfield[2]{#2}
\providecommand\bibinfo[2]{#2}
\providecommand\natexlab[1]{#1}
\providecommand\showeprint[2][]{arXiv:#2}

\bibitem[Agarwal et~al\mbox{.}(2024)]%
        {agarwal2024gkd}
\bibfield{author}{\bibinfo{person}{Rishabh Agarwal}, \bibinfo{person}{Nino
  Vieillard}, \bibinfo{person}{Yongchao Zhou}, \bibinfo{person}{Piotr
  Stanczyk}, \bibinfo{person}{Sabela Ramos}, \bibinfo{person}{Matthieu Geist},
  {and} \bibinfo{person}{Olivier Bachem}.} \bibinfo{year}{2024}\natexlab{}.
\newblock \showarticletitle{On-Policy Distillation of Language Models: Learning
  from Self-Generated Mistakes}. In \bibinfo{booktitle}{\emph{International
  Conference on Learning Representations (ICLR)}}.
\newblock
\showeprint[arxiv]{2306.13649}


\bibitem[Chen et~al\mbox{.}(2025)]%
        {chen2025retaining}
\bibfield{author}{\bibinfo{person}{Howard Chen}, \bibinfo{person}{Noam Razin},
  \bibinfo{person}{Karthik Narasimhan}, {and} \bibinfo{person}{Danqi Chen}.}
  \bibinfo{year}{2025}\natexlab{}.
\newblock \bibinfo{title}{Retaining by Doing: The Role of On-Policy Data in
  Mitigating Forgetting}.
\newblock
\showeprint[arxiv]{2510.18874}


\bibitem[Chen et~al\mbox{.}(2021)]%
        {chen2021finqa}
\bibfield{author}{\bibinfo{person}{Zhiyu Chen}, \bibinfo{person}{Wenhu Chen},
  \bibinfo{person}{Charese Smiley}, \bibinfo{person}{Sameena Shah},
  \bibinfo{person}{Iana Borova}, \bibinfo{person}{Dylan Langdon},
  \bibinfo{person}{Reema Moussa}, \bibinfo{person}{Matt Beane},
  \bibinfo{person}{Ting-Hao Huang}, \bibinfo{person}{Bryan Routledge}, {and}
  \bibinfo{person}{William~Yang Wang}.} \bibinfo{year}{2021}\natexlab{}.
\newblock \showarticletitle{FinQA: A Dataset of Numerical Reasoning over
  Financial Data}. In \bibinfo{booktitle}{\emph{Proceedings of the 2021
  Conference on Empirical Methods in Natural Language Processing (EMNLP)}}.
  \bibinfo{pages}{3697--3711}.
\newblock
\showeprint[arxiv]{2109.00122}


\bibitem[Gu et~al\mbox{.}(2024)]%
        {gu2024minillm}
\bibfield{author}{\bibinfo{person}{Yuxian Gu}, \bibinfo{person}{Li Dong},
  \bibinfo{person}{Furu Wei}, {and} \bibinfo{person}{Minlie Huang}.}
  \bibinfo{year}{2024}\natexlab{}.
\newblock \showarticletitle{MiniLLM: Knowledge Distillation of Large Language
  Models}. In \bibinfo{booktitle}{\emph{International Conference on Learning
  Representations (ICLR)}}.
\newblock
\showeprint[arxiv]{2306.08543}


\bibitem[Hinton et~al\mbox{.}(2015)]%
        {hinton2015distilling}
\bibfield{author}{\bibinfo{person}{Geoffrey Hinton}, \bibinfo{person}{Oriol
  Vinyals}, {and} \bibinfo{person}{Jeff Dean}.}
  \bibinfo{year}{2015}\natexlab{}.
\newblock \bibinfo{title}{Distilling the Knowledge in a Neural Network}.
\newblock
\showeprint[arxiv]{1503.02531}


\bibitem[Hu et~al\mbox{.}(2025)]%
        {hu2025fintrust}
\bibfield{author}{\bibinfo{person}{Tiansheng Hu}, \bibinfo{person}{Tongyan Hu},
  \bibinfo{person}{Liuyang Bai}, \bibinfo{person}{Yilun Zhao},
  \bibinfo{person}{Arman Cohan}, {and} \bibinfo{person}{Chen Zhao}.}
  \bibinfo{year}{2025}\natexlab{}.
\newblock \showarticletitle{FinTrust: A Comprehensive Benchmark of
  Trustworthiness Evaluation in Finance Domain}. In
  \bibinfo{booktitle}{\emph{Proceedings of the 2025 Conference on Empirical
  Methods in Natural Language Processing (EMNLP)}}.
  \bibinfo{pages}{10099--10128}.
\newblock
\showeprint[arxiv]{2510.15232}


\bibitem[Islam et~al\mbox{.}(2023)]%
        {islam2023financebench}
\bibfield{author}{\bibinfo{person}{Pranab Islam}, \bibinfo{person}{Anand
  Kannappan}, \bibinfo{person}{Douwe Kiela}, \bibinfo{person}{Rebecca Qian},
  \bibinfo{person}{Nino Scherrer}, {and} \bibinfo{person}{Bertie Vidgen}.}
  \bibinfo{year}{2023}\natexlab{}.
\newblock \bibinfo{title}{FinanceBench: A New Benchmark for Financial Question
  Answering}.
\newblock
\showeprint[arxiv]{2311.11944}


\bibitem[Kim and Rush(2016)]%
        {kim2016seqkd}
\bibfield{author}{\bibinfo{person}{Yoon Kim} {and}
  \bibinfo{person}{Alexander~M. Rush}.} \bibinfo{year}{2016}\natexlab{}.
\newblock \showarticletitle{Sequence-Level Knowledge Distillation}. In
  \bibinfo{booktitle}{\emph{Proceedings of the 2016 Conference on Empirical
  Methods in Natural Language Processing (EMNLP)}}.
  \bibinfo{pages}{1317--1327}.
\newblock
\showeprint[arxiv]{1606.07947}


\bibitem[Lu and {Thinking Machines Lab}(2025)]%
        {lu2025opd}
\bibfield{author}{\bibinfo{person}{Kevin Lu} {and} \bibinfo{person}{{Thinking
  Machines Lab}}.} \bibinfo{year}{2025}\natexlab{}.
\newblock \bibinfo{title}{On-Policy Distillation}.
\newblock \bibinfo{howpublished}{Thinking Machines Lab: Connectionism}.
\newblock
\href{https://doi.org/10.64434/tml.20251026}{doi:\nolinkurl{10.64434/tml.20251026}}


\bibitem[{OpenAI}(2025)]%
        {openai2025gptoss}
\bibfield{author}{\bibinfo{person}{{OpenAI}}.} \bibinfo{year}{2025}\natexlab{}.
\newblock \bibinfo{title}{gpt-oss-120b \& gpt-oss-20b Model Card}.
\newblock
\showeprint[arxiv]{2508.10925}


\bibitem[Snell et~al\mbox{.}(2022)]%
        {snell2022context}
\bibfield{author}{\bibinfo{person}{Charlie Snell}, \bibinfo{person}{Dan Klein},
  {and} \bibinfo{person}{Ruiqi Zhong}.} \bibinfo{year}{2022}\natexlab{}.
\newblock \bibinfo{title}{Learning by Distilling Context}.
\newblock
\showeprint[arxiv]{2209.15189}


\bibitem[Tang et~al\mbox{.}(2025)]%
        {tang2025financereasoning}
\bibfield{author}{\bibinfo{person}{Zichen Tang}, \bibinfo{person}{Haihong E},
  \bibinfo{person}{Ziyan Ma}, \bibinfo{person}{Haoyang He},
  \bibinfo{person}{Jiacheng Liu}, \bibinfo{person}{Zhongjun Yang},
  \bibinfo{person}{Zihua Rong}, \bibinfo{person}{Rongjin Li},
  \bibinfo{person}{Kun Ji}, \bibinfo{person}{Qing Huang},
  \bibinfo{person}{Xinyang Hu}, \bibinfo{person}{Yang Liu}, {and}
  \bibinfo{person}{Qianhe Zheng}.} \bibinfo{year}{2025}\natexlab{}.
\newblock \showarticletitle{FinanceReasoning: Benchmarking Financial Numerical
  Reasoning More Credible, Comprehensive and Challenging}. In
  \bibinfo{booktitle}{\emph{Proceedings of the 63rd Annual Meeting of the
  Association for Computational Linguistics (ACL)}}.
  \bibinfo{pages}{15748--15769}.
\newblock
\showeprint[arxiv]{2506.05828}


\bibitem[Wang et~al\mbox{.}(2024)]%
        {wang2024mmlupro}
\bibfield{author}{\bibinfo{person}{Yubo Wang}, \bibinfo{person}{Xueguang Ma},
  \bibinfo{person}{Ge Zhang}, \bibinfo{person}{Yuansheng Ni},
  \bibinfo{person}{Abhranil Chandra}, \bibinfo{person}{Shiguang Guo},
  \bibinfo{person}{Weiming Ren}, \bibinfo{person}{Aaran Arulraj},
  \bibinfo{person}{Xuan He}, \bibinfo{person}{Ziyan Jiang},
  \bibinfo{person}{Tianle Li}, \bibinfo{person}{Max Ku}, \bibinfo{person}{Kai
  Wang}, \bibinfo{person}{Alex Zhuang}, \bibinfo{person}{Rongqi Fan},
  \bibinfo{person}{Xiang Yue}, {and} \bibinfo{person}{Wenhu Chen}.}
  \bibinfo{year}{2024}\natexlab{}.
\newblock \showarticletitle{MMLU-Pro: A More Robust and Challenging Multi-Task
  Language Understanding Benchmark}. In \bibinfo{booktitle}{\emph{Advances in
  Neural Information Processing Systems 37 (NeurIPS), Datasets and Benchmarks
  Track}}. \bibinfo{pages}{95266--95290}.
\newblock
\showeprint[arxiv]{2406.01574}


\bibitem[Yeo et~al\mbox{.}(2026)]%
        {yeo2026hintsd}
\bibfield{author}{\bibinfo{person}{Woongyeong Yeo}, \bibinfo{person}{Yumin
  Choi}, \bibinfo{person}{Taekyung Ki}, {and} \bibinfo{person}{Sung~Ju Hwang}.}
  \bibinfo{year}{2026}\natexlab{}.
\newblock \bibinfo{title}{HINT-SD: Targeted Hindsight Self-Distillation for
  Long-Horizon Agents}.
\newblock
\showeprint[arxiv]{2605.17873}


\bibitem[Zhao et~al\mbox{.}(2026)]%
        {zhao2026opsd}
\bibfield{author}{\bibinfo{person}{Siyan Zhao}, \bibinfo{person}{Zhihui Xie},
  \bibinfo{person}{Mengchen Liu}, \bibinfo{person}{Jing Huang},
  \bibinfo{person}{Guan Pang}, \bibinfo{person}{Feiyu Chen}, {and}
  \bibinfo{person}{Aditya Grover}.} \bibinfo{year}{2026}\natexlab{}.
\newblock \bibinfo{title}{Self-Distilled Reasoner: On-Policy Self-Distillation
  for Large Language Models}.
\newblock
\showeprint[arxiv]{2601.18734}


\bibitem[Zhu et~al\mbox{.}(2021)]%
        {zhu2021tatqa}
\bibfield{author}{\bibinfo{person}{Fengbin Zhu}, \bibinfo{person}{Wenqiang
  Lei}, \bibinfo{person}{Youcheng Huang}, \bibinfo{person}{Chao Wang},
  \bibinfo{person}{Shuo Zhang}, \bibinfo{person}{Jiancheng Lv},
  \bibinfo{person}{Fuli Feng}, {and} \bibinfo{person}{Tat-Seng Chua}.}
  \bibinfo{year}{2021}\natexlab{}.
\newblock \showarticletitle{TAT-QA: A Question Answering Benchmark on a Hybrid
  of Tabular and Textual Content in Finance}. In
  \bibinfo{booktitle}{\emph{Proceedings of the 59th Annual Meeting of the
  Association for Computational Linguistics (ACL)}}.
  \bibinfo{pages}{3277--3287}.
\newblock
\showeprint[arxiv]{2105.07624}


\end{thebibliography}

\end{document}